\documentclass[review]{elsarticle}
\usepackage{float}
\pdfoutput=1 
\usepackage{hyperref}
\usepackage{url}

\usepackage{amsfonts,amssymb}
\usepackage{amsmath}
\usepackage{amssymb}
\usepackage{caption}
\usepackage{algorithm}
\usepackage{bm}
\usepackage{amsmath,amssymb,amsfonts}
\usepackage{algorithmic}
\usepackage{graphicx}
\usepackage{textcomp}
\usepackage{subfigure}
\usepackage{makecell}
\usepackage{balance}
\usepackage{multirow}
\usepackage{titlesec}
\numberwithin{equation}{section}
\numberwithin{figure}{section}
\numberwithin{table}{section}

\begin{document}

\captionsetup[figure]{labelfont={bf},labelformat={default},labelsep=period,name={Fig.}}
\captionsetup[table]{labelfont={bf},labelformat={default},labelsep=period,name={Table.}}
\begin{frontmatter}

\title{Person Re-identification based on Robust Features in \\ Open-world\tnoteref{mytitlenote}}

\author[mymainaddress]{Yaguan Qian}

\author[mymainaddress]{Anlin Sun}
\author[mysecondaryaddress]{Xiaohui Guan}
\author[mymainaddress]{Qiong Lou\corref{mycorrespondingauthor}}
\cortext[mycorrespondingauthor]{Corresponding author}
\ead{bearqiong@163.com}
\author[mymainaddress]{Wei Li}
\address[mymainaddress]{Laboratory of Artificial Intelligence, School of Science, Zhejiang University of Science and Technology, Hangzhou 310012, China}
\address[mysecondaryaddress]{College of Information Engineering, Zhejiang University of Water Resources and Electric Power, Hangzhou 310018, China}

\begin{abstract}
 Deep learning technology promotes the rapid development of person re-identifica-tion (re-ID). However, some challenges are still existing in the open-world. First, the existing re-ID research usually assumes only one factor variable (view, clothing, pedestrian pose, pedestrian occlusion, image resolution, RGB/IR modality) changing, ignoring the complexity of multi-factor variables in the open-world. Second, the existing re-ID methods are over depend on clothing color and other apparent features of pedestrian, which are easily disguised or changed. In addition, the lack of benchmark datasets containing multi-factor variables is also hindering the practically application of re-ID in the open-world. In this paper, we propose a low-cost and high-efficiency method to solve shortcomings of the existing re-ID research, such as unreliable feature selection, low efficiency of feature extraction, single research variable, etc. Our approach based on pose estimation model improved by group convolution to obtain the continuous key points of pedestrian, and utilize dynamic time warping ($DTW$) to measure the similarity of features between different pedestrians. At the same time, to verify the effectiveness of our method, we provide a miniature dataset which is closer to the real world and includes pedestrian changing clothes and cross-modality factor variables fusion. Extensive experiments are conducted and the results show that our method achieves Rank-1: 60.9\%, Rank-5: 78.1\%, and mAP: 49.2\% on this dataset, which exceeds most existing state-of-art re-ID models.
\end{abstract}


\end{frontmatter}

\section{Introduction}

\textbf{Person re-identification (re-ID)} is an important branch of computer vision, which achieves the cross-camera tracking of the target person with a breakthrough of time and space limitation of single camera, with broad prospects in the field of public security, Intelligent security, etc. Great progress has been made in research in re-ID due to the development of deep learning techniques, presenting results that can perform relatively high recognition accuracy in public dataset\cite{label1}. However, most of those are results of re-ID in close-world based on the hypothesis of only one factor variable (view, clothing, pedestrian pose, pedestrian occlusion, image resolution, RGB/IR modality) changed. Plenty of difficulties and challenges occur in the transition of re-ID research from closed-world to open-world owing to the complexity of multi-factor variables in open-world\cite{label2}. Specifically, related difficulties and challenges\cite{label15,label16,label19} include: i) lacking of the re-ID benchmark datasets which containing multi-factor variables has led to existing re-ID research centering on only a few factor variables; ii) Existing re-ID research rely on unreliable pedestrian apparent features that be easily disguised or changed; iii) Existing re-ID models are inefficient in extracting pedestrian features from images and are easily effected by images factors (resolution ratio, color, light, angle, etc.). 

Due to the lack of benchmark datasets of re-ID in open-world, re-ID research based on deep learning can only focus on a few factor variables and cannot cope with the situation of multi-factor changes in open-world. The existing re-ID benchmark datasets belong to closed-world, most of which only contain single-factor variable. For instance, the common benchmark datasets such as ReID – Market1501\cite{label3}, CUHK03\cite{label4}, DukeMTMC – ReID\cite{label5} only contain the view factor. It is considered effective to create open-world re-ID dataset with multi-factor variables to improve the usability of re-ID deep learning model in open-world. However, the collection of dataset is difficult since it not only consumes considerable manpower and resources, but also involve sensitive privacy issues, which makes it hard to build a dataset in most of the countries that pay high attention to citizen privacy security. Therefore, we use other methods to search for general pedestrian robust features instead of training re-ID deep learning model. Then, we transform those general pedestrian robust features into reliable pedestrian identity information that we can rely on to re-identify this specific pedestrian. 

Basically, a person's identity should be decided by his biological features (lineament, height, etc.), but not his unreliable apparent features (clothes, shoes, haircuts, etc.)\cite{label18}. However, the pedestrian identification based on existing re-ID models highly relies on such unreliable apparent information which can easily be artificially changed or disguised on purpose, leading to a low usability of re-ID research in open-world. It is always a difficult task to extract biological features of pedestrian, since there is a high relativity between the quality of those extracted features and the factors of an image including resolution ratio, light, angle, color, not to mention the artificial disguise of features. Thus what matters more is how to search for an effective method to extract some stable biological features from images. What we considered is a pedestrian’s gait as biological features that is almost fixed in an uninterrupted period of time\cite{label31}.  At the meantime, since the pose estimation model\cite{label6,label32} is robust to image resolution, color, view and other open-world variables, we extract pedestrian skeleton key points as general robust feature information based on pose estimation. Moreover, the continuous frame images can transform the general robust feature information of pedestrians into reliable identity information which has advantages of long distance and longtime effectiveness\cite{label31}.

Our method is presented under the circumstances of multi-factor variables in open-world. We use the pose estimation model to extract the continuous feature information of pedestrian skeleton key points based on video continuous frames instead of using unreliable apparent feature information of pedestrian during cross-camera tracking, those continuous feature information of pedestrian skeleton key points is specific identity information of pedestrian. Then according to the similarities of matching each pedestrian’s identity information to re-identify the specific pedestrian. Compared with existing re-ID models, there are three main advantages of our method. Firstly, the continuous feature information of pedestrian skeleton key points extracted by pose estimation model is robust and not easy to be disguised or changed. Secondly, our method is more efficient in extracting pedestrian features owing to the insensitivity of the pose estimation model to image resolution, color, view and other factors. Lastly, we consider the distance between feature vectors as optimization problem, and slove it based on dynamic programming, which makes our results more accurate. These above make our method improve the usability of re-ID in open-world in the absence of relevant datasets. 

Our study is based on assumptions where we analyze and sum up existing re-ID research, then propose a method combining deep learning with traditional algorithms to solve the problem of low availability of re-ID in open-world. We proved the effectiveness of our methodology based on experiment under the condition of pedestrian changing clothes and cross-modality, then compared it with existing state-of-art re-ID research. At the same time, our approach overcomes the challenge of lacking relevant dataset. The contribution of this paper can be summarized as follows:
\begin{enumerate}[(1)]
\item We reveal the poor usability of the existing re-ID research in open-world, and the fact that existing re-ID research are derailed from the real world. Moreover, we further demonstrate the disadvantages of the existing re-ID on the basis of previous research.
\item Combined with previous research, we optimized $DTW$ algorithm by using global constrains, lower-bounded function, early stop and other optimization strategies. We also analyzed the computational complexity of various optimization combinations, which can provide reference for related research. Moreover we improved the pose estimation model HigherHRNet by replacing a part of residual blocks with ResNeXt blocks, reducing the space complexity of the algorithm. 
\item We presented a micro dual-factor re-ID dataset based on video frame sequence including pedestrian changing clothes and cross-modality. The total parameters of this modified network decreased by 15\%. The number of FLOPs of forward reasoning decreased by 5\%. The network performance was improved by 0.5\%, with average accuracy(AP) reaching 67.9\% and 66.9\% on COCO2017 verification dataset and test dataset respectively. 
\end{enumerate}

\section{Related Work}

Some existing re-ID models have been used in open-world, however, the performance of these models needs to be further improved in open-world. According to the types of features that extracted by the model, these models are divided into two categories, based on the apparent features and the biological features of pedestrians.

\textbf{Research Based on Apparent Features.} The DG-Net\cite{label1} model contains a generation module, which encodes each pedestrian into appearance code and structure code, and then exchanges appearance code and structure code between different pedestrians, so that the model can learn different attributes of the same pedestrian features to improve the performance of the model. Although the DG-Net model learned different attributes(appearance and structure features) of the same pedestrian during training, it still cannot cope with the multi-factor variables change in open-world. There are two main reasons: i) The attribute features, such as facial features, learned by the model are unstable and easily disguised. ii) The model has low efficiency in extracting pedestrian attribute feature information from the image and cannot counter the interference of factors such as image resolution. Similarly, the ReIDCaps\cite{label9} model has the same disadvantage. Baseline of AGW\cite{label17} can better adapt to single-/cross-modality, but it does not involve the impact of other factors on the performance of the model, and only uses the image information of a single frame.

\textbf{Research Based on Biological Features.} \cite{label18} propose an idea using radio signals to collect pedestrian feature information as a basis for re-ID. Although this collected information belongs to biological features of pedestrian, it is also unstable and easily disguised artificially, such as carrying interference sources. In addition, special equipment is required to transmit or receive radio signals, which makes it unrealistic to equip on a large scale. Research based on depth maps\cite{label10,label11,label12,label13,label14} can capture reliable identity information of pedestrians, alone with a certain anti-interference ability against external factors, in theory, which can promote the implementation of re-ID in open-world. However, special equipment is required for capturing images with depth information, making it unsuitable for large-scale deployment. Some recent research have taken a new perspective to extract pedestrian contour features as a basis for re-identification\cite{label15,label16}. However, regret to say that these research cannot promote the application of re-ID in open-world, for two reasons: i) Pedestrian contour features are easily changed with changes in clothing, making it lack of a long-term effects. ii) Even without the influence of external factors, it can still be difficult to accurately extract the contours of pedestrians especially in some complex occasions such as covering, clothing or pedestrians traveling at night.

\textbf{Similarities and differences with gait recognition research.} Gait recognition\cite{label31,label33,label34}, a relatively new research field, has some similarities and differences compared with our research. Different from the existing gait recognition research, our method benefits from the pose estimation model, which can obtain more stable and reliable feature information of pedestrian skeleton key points from images, without the usage of segmentation or spatial subtraction to obtain the pedestrian's unstable body contour features. and our method has more relaxed constraints without following assumptions: i) The silhouette of the pedestrian should be segmented from the background. ii) Pedestrians’ body contours will not change after cloth-changing. iii) Pedestrian pixels should not be too few in the image. On the other hand, what our research is similar to the existing gait recognition research is that both research are based on continuous frames of video with the usage of the related information between each frames.

\section{Problem Description}

Most of the existing re-ID research are under the condition of close-world, assuming that only one factor variable (view, clothing, pedestrian pose, pedestrian occlusion, image resolution, RGB/IR modality) changing, which is far from reality. Research in open-world faces greater difficulties and challenges due to the complexity of the scene. From the perspective of probability theory, we can simply describe the difference between re-ID in close-world and open-world. Define the set of open-world factor variables as $\Omega=\{v_{1},v_{2},v_{3},\cdots,v_{i}\}$, $i\in\{1,2,3,\cdots\}$, Event $A_{i}$ represents being affected by the factor $v_{i}$ in $\Omega$ during pedestrian retrieval, and event $\delta$ represents successfully retrieve a pedestrian across the camera. Then, re-ID in close-world can be expressed as a conditional probability $P(\delta\;|\;A_{i})$, while it can be expressed as $P(\delta\;|\;A_{1}\cap A_{2}\cap A_{3}\cdots \cap A_{i})$ in open-world .

We have revealed in the previous part that the existing re-ID model is extremely dependent on apparent information of pedestrian such as color of cloth when judging the identity of pedestrians under the situation that this information is extremely unreliable under the background of multi-factor changes in open-world\cite{label15,label16,label19}. In order to further illustrate the dependence of the re-ID model in close-world on the color of pedestrian clothing, we take the state-of-art model DG-Net, which has learned multiple feature (appearance and structure) information, as an example. We remove the final classification layer of DG-Net, output the 1024-dimensional feature vectors before and after the change of the same pedestrian on the testset PRCC in \cite{label15}, then calculate the cosine similarity of these feature vectors. Result is shown in Fig. 3.1. Generally, the cosine similarity of the feature vectors is low with the highest value less than 0.4. In the 1024-dimensional space, there is a relatively large directional deviation between the eigenvectors while showing a reverse trend. Though the DG-Net model learns the attribute information of pedestrians during training, it still relies heavily on apparent information of pedestrian such as the colors of clothing.

\begin{figure}
\setlength{\abovecaptionskip}{-0.2cm}   
\setlength{\belowcaptionskip}{-0.5cm}    
\begin{center}
\includegraphics[width=9.8cm,height=4.1cm]{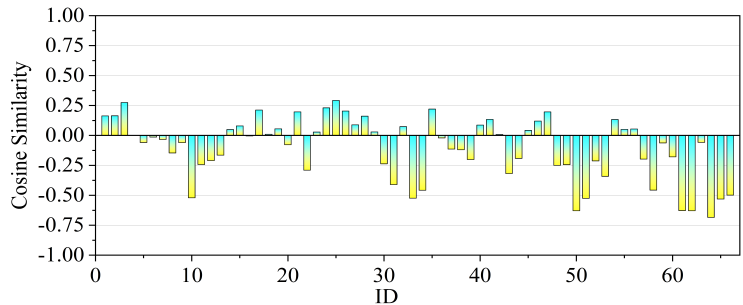} 
\end{center}
\caption{The test result of DG-Net on the PRCC testset. The ordinate is the cosine similarity between the feature vectors before and after the change of each pedestrian with an average value of -0.215; the abscissa is the ID of each pedestrian; PRCC is the pedestrian refitting dataset containing 71 pedestrians from where we selected 67; the DG-Net model is taken from the official open source website.}
\label{figa14}
\end{figure}

Here comes the question: how to weaken the dependence of the re-ID model on the apparent information of pedestrians when judging the identity of pedestrians? An effective method is to train a deep learning model based on the re-ID dataset of multi-factor in open-world to make this model robust to multi-factor variables, but it is unrealistic. Therefore, instead of directly solving this problem, we use the idea of limitation to further abstract this problem. Thus, this issue can be transformed into searching a way that the re-ID model judge the identity of the pedestrians without relying on their apparent information.

\section{Methodology}

Specifically, in our method, continuous frame images are used as the input of pose estimation model, while the time series features of the pedestrian skeleton key points are obtained and used as the basis of re-identification. It’s a common case that the lengths of two time series features are unequal, since the length depends on the number of video frames. Therefore, we consider this distance measurement of time series features as an optimization problem. Based on the idea of dynamic programming, we use dynamic time warping ($DTW$) to solve the distance between time series features, then select the features closest to the model feature as query results, so as to complete the matching of pedestrians with the same identity. 

Good models (or methods) should occupy as little computing resource or storage space as possible. It is inevitable that there will be computational redundancy since our method includes both traditional algorithms and deep learning techniques. According to the characteristics of our method, we divide it into two stages: stage.1 based on deep learning and stage.2 based on traditional algorithm and use corresponding optimization strategies for these two stages to reduce the computational complexity of the algorithm. Since the detection accuracy of skeleton key points will affect the subsequent feature matching, and considering that there are few pixels of the person in the surveillance video, the HigherHRNet\cite{label6} network with ultra-high detection accuracy and high-resolution feature maps is considered to be an appropriate pose estimation model. However, the HigherHRNet network has a large number of parameters, with more than 300 convolutional layers, and numerous stacked residual blocks. \cite{label7} shows that the ResNeXt block is better than the residual block, and the group convolution has a smaller number of parameters. Thus we replace some of the residual blocks in the network with the ResNeXt block using group convolution, and the improved network is denoted as HigherHRNetX. Stage.2 regarding the distance solution between time series features as an optimization problem will undoubtedly increase the computational complexity of the algorithm. In view of the characteristics of $DTW$, we use optimization strategies such as global constraints, lower-bounded functions, and early stop to reduce the complexity, and quantitatively analyze the effects of each strategy.

Denote HigherHRNetXt as a function $G(x)$, the output as a time series sequence $Y$. The obtained continuous frames information of pedestrian $S$ to be required at place $A$ is defined as $X_{s}(t)=\{X_{1},X_{2},X_{3},\cdots,X_{t}\}$, $t\geq C$; the information of suspected pedestrian $f$ at place $B$ is defined as $Z_{f}(t)=\{Z_{1},Z_{2},Z_{3},\cdots,Z_{t}\}$, $t\geq C$. Obtain the time series features of skeletons key points of $S$ and $f$, as $Y_{A}=G(X_{s}(t))$, $Y_{B}=G(Z_{f}(t))$. Thus, the distance between $Y_{A}$ and $Y_{B}$ can be written as the following function: $D(Y_{A},Y_{B})$. The lengths of time series features $Y_{A}$ and $Y_{B}$ are $l_{A}, l_{B}\in[C,+\infty]$, respectively. Consider the solution of the distance between $Y_{A}$ and $Y_{B}$ as an optimization problem, which is the following constrained minimization problem :
\begin{equation}\begin{array}{l}
\min \quad D\left(Y_{A}, Y_{B}\right) \\
\; \text {s.t.\quad} \; \, l_{A} \geq C,\\
\quad \quad \quad l_{B}\geq C.
\end{array}
\end{equation}

Eq.(4.1) can be solved based on the $DTW$\cite{label20} algorithm to obtain the best warping path of $Y_{A}$ and $Y_{B}$. Denote the minimum matching distance as $d\in[0,+\infty]$. The shorter the distance $d$, the higher the similarity between $Y_{A}$ and $Y_{B}$, and the greater the probability value $P(f=S)$; conversely, the lower the similarity, the smaller the probability value $P(f=S)$.

\subsection{Assumptions}
Our research focus on the image field, that is, only extracting information from the image. Obviously, more image information brings more benefit to our research. At the same time, the open-world contains many factor variables. Each additional factor variable in the research process of re-ID will make the research more difficult. At present, no specific way to deal with all the open-world factor variables has been found in our research. So we need to establish the following assumptions:

\textbf{Assumption 1.} The feature information obtained by the re-ID model is only obtained from the image, and does not contain auxiliary information such as text.

\textbf{Assumption 2.} Suppose the pedestrian to be retrieved is $S$, and the query image about $S$ is a continuous frame, that is, $X_{s}(t)=\{X_{1},X_{2},X_{3},\cdots,X_{t}\},t\geq C$. $X_{t}$ is the $t^{th}$ frame of $S$, $X_{s}(t)$ is the initial query information of $S$, and $C$ is the video frame rate.

\textbf{Assumption 3.} Denote pedestrian clothing as factor $v_{1}$, image cross-modality as factor $v_{2}$, \{$v_{1},v_{2}\}\in \Omega$. We accomplish our research on $v_{1}$ and $v_{2}$.

Assumption 1 makes the problem in Part.3 equivalent to how to make the re-ID model use the pedestrian biological feature information in the image to identify pedestrians. At the same time, on the basis of Assumption 1, 2 and 3, our research problem is determined as 'under the influence of  pedestrian changing clothes, cross-modality, based on continuous frames of video, how the re-ID study extract the biological features of pedestrians in the image and complete the recognition of specific pedestrians,' that is 
$\max\{P(\delta\;|\;A_{1}\cap A_{2})\}$.

\subsection{HigherHRNetXt Network}
\begin{figure}[H]
\setlength{\abovecaptionskip}{-0.2cm}   
\setlength{\belowcaptionskip}{-0.5cm}    
\begin{center}
\includegraphics[width=8.5cm,height=4.5cm]{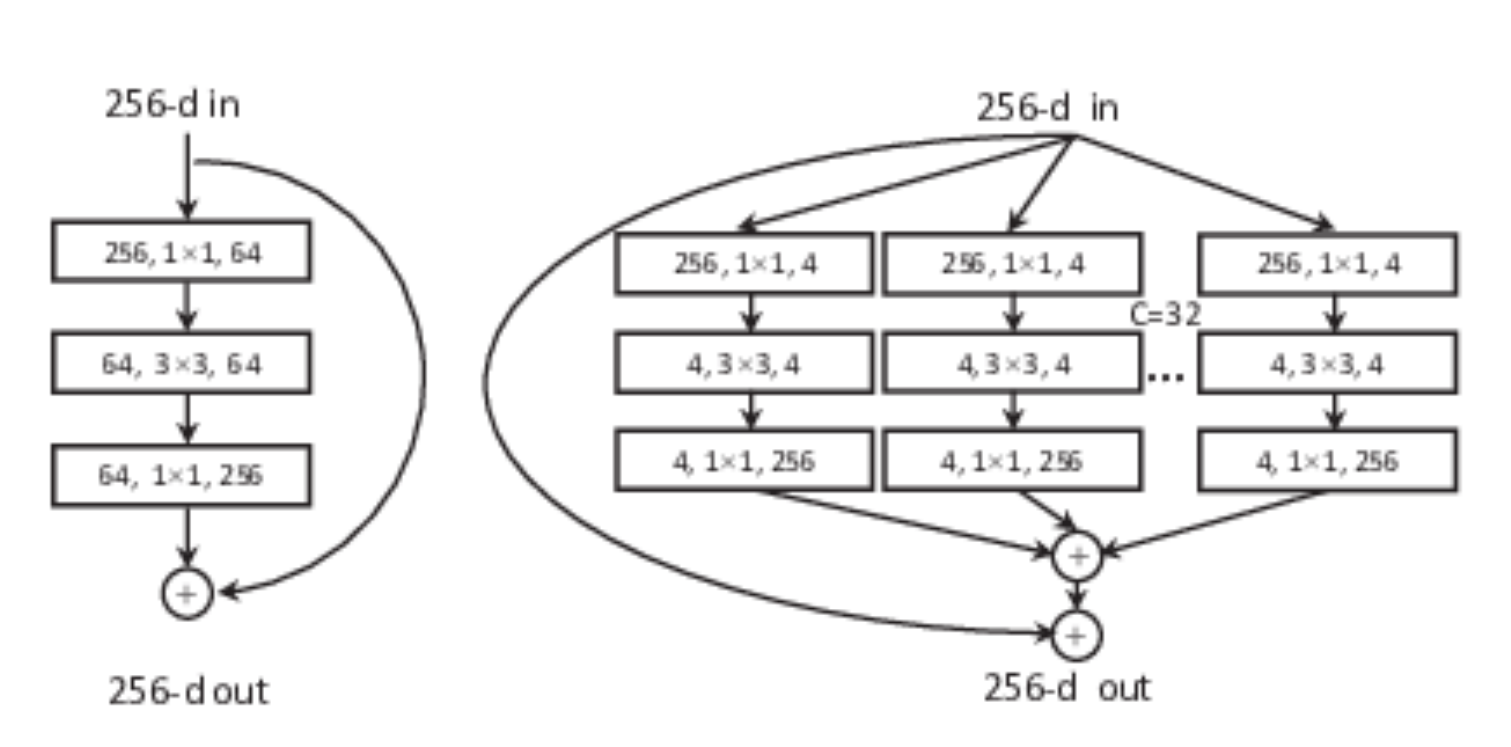} 
\end{center}
\caption{On the left is a kind of ResNet block, the input and output dimension are both 256; On the right is one of ResNeXt block, the base input dimension is 256, the output dimension is 256, the base C is 32.}
\label{figa7}
\end{figure}
\textbf{ResNeXt block.} As an improvement to the network residual block convolution structure, ResNeXt uses a group convolution method to make the convolutional network improve the performance and reduce the total number of network parameters without increasing the depth and width, but only changing the base. $F(X)=\sum_{i=1}^{C} T_{i}(x)$, $C$ is the base, which represents the number of identical branches in a module; $T_{i}(x)$ is the $i^{th}$ branch of the same topology. Fig. 4.1 illustrates the architecture comparison between residual block and ResNeXt block.

\textbf{HigherHRNetXt Structure.} Replace part of the residual blocks in HigherHRNet with ResNeXt blocks, the network structure is shown in Fig. 4.2.
\begin{figure}[H]
\setlength{\abovecaptionskip}{-0.2cm}   
\setlength{\belowcaptionskip}{-0.5cm}    
\begin{center}
\includegraphics[width=12cm,height=8cm]{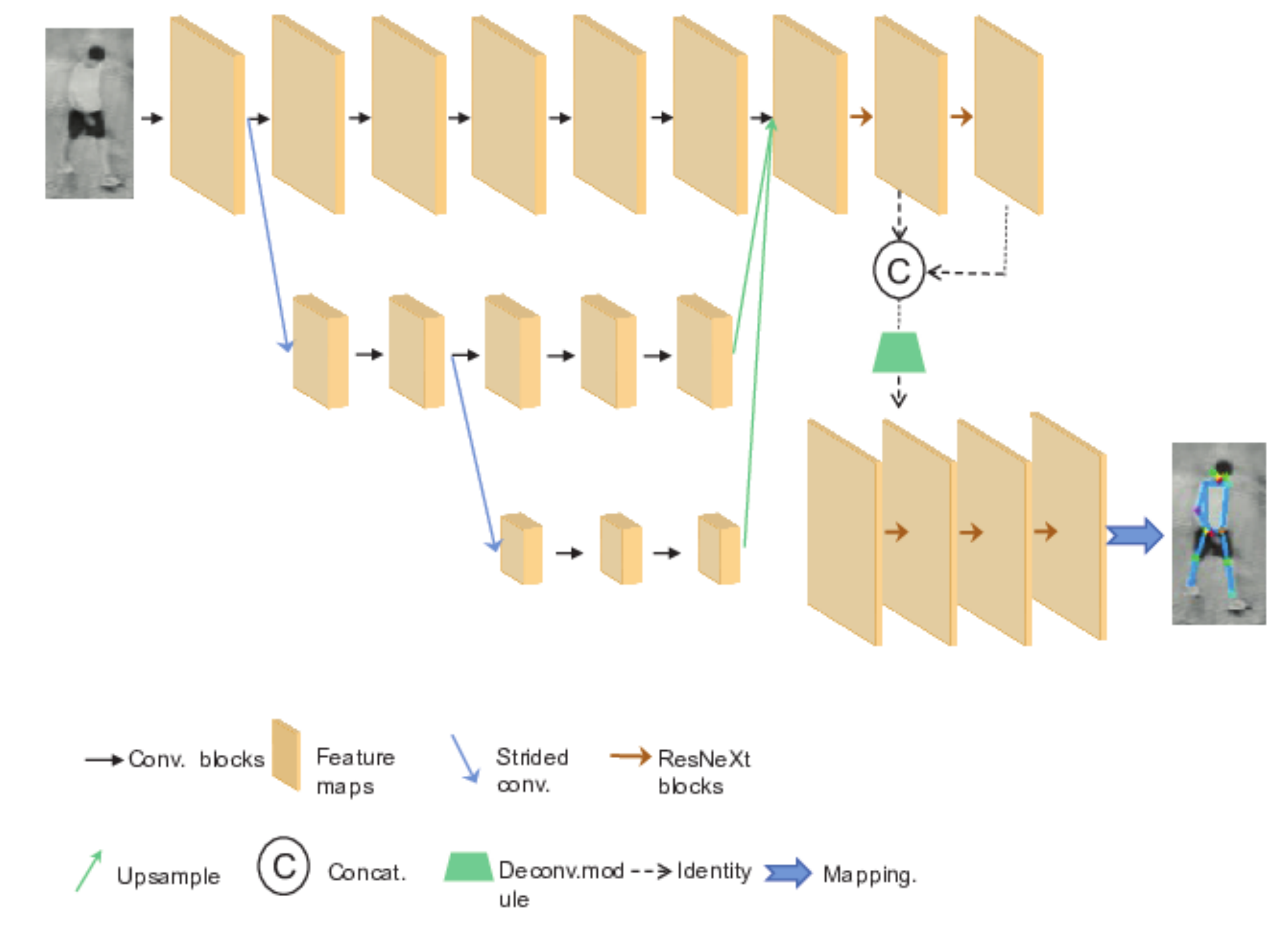} 
\end{center}
\caption{HigherHRNetXt network structure.}
\label{figa8}
\end{figure}

\subsection{Dynamic Time Warping ($DTW$)}{}
$DTW$ algorithm, first proposed by Japanese scholar Itakura, uses the idea of dynamic programming to solve the minimum distance between two time series $x_{1}$ and $x_{2}$ with equal or unequal length, and has high accuracy in measuring the similarity of time series.  $DTW$ algorithm is widely applied and there are many related research. In order to improve the computational efficiency of $DTW$ while reducing computational cost, we set some constraints on $DTW$. We will introduce these constraints and analyze the computational complexity of each optimization strategy.
\subsubsection{constraint conditions}
\begin{figure}[H]
\setlength{\abovecaptionskip}{-0.2cm}   
\setlength{\belowcaptionskip}{-0.5cm}    
\begin{center}
\includegraphics[width=7cm,height=5.5cm]{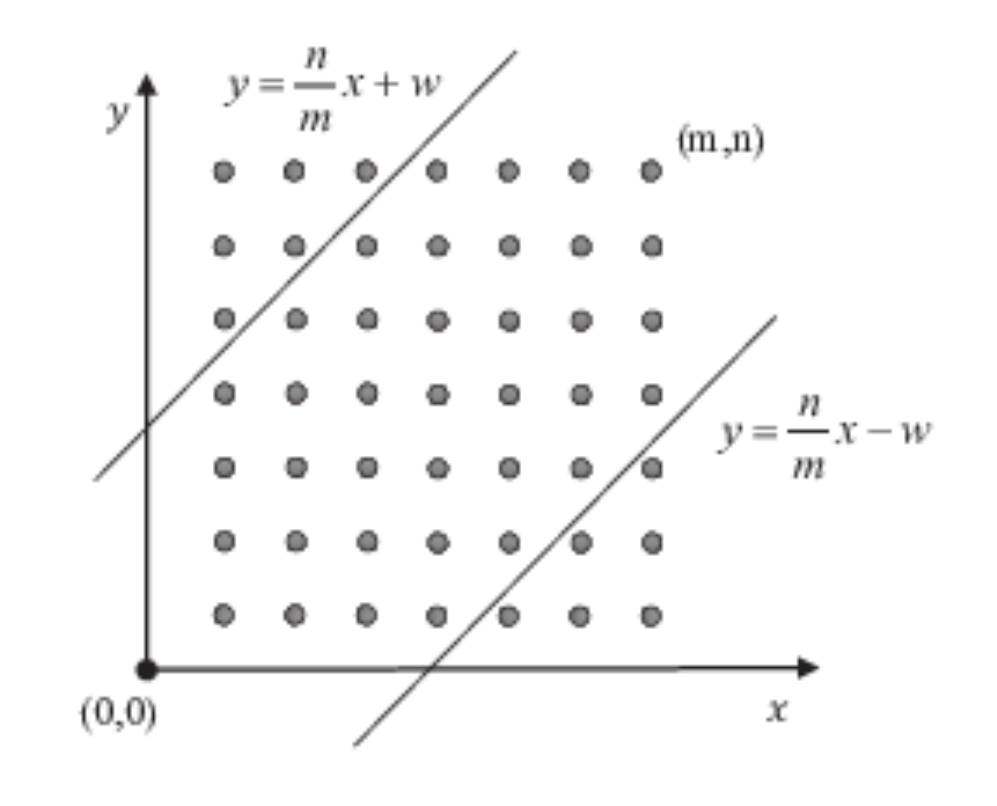} 
\end{center}
\caption{Gray dot represents the elements in the matrix $M$.}
\label{figa9}
\end{figure}
\textbf{Global Constraints.} Constraining the search space of $DTW$ can reduce the number of elements to be calculated, and reduce the time complexity of the algorithm. Therefore, we set a boundary line in the $DTW$ search space, that is, the Sakoe-Chiba\cite{label21} constraint. The location of the search boundary is determined by the width $w$ of the warping window. $w=r \times \max (m, n), r \in[0,1]$,$m$,$n$ is the length of the sequence. If the set of euclidean distances of each element in time series $Q$ and $L$ is a matrix $M$, the constraint can be visualized by projecting $M$ into Cartesian coordinate, as shown in Fig. 4.3. The constrained $DTW$ reduces the time complexity of the algorithm due to the reduction of its search space, at the same time avoids excessive curvature of the warping path. The warping window $\chi$ of $DTW$ is as follows:
\begin{equation}
\chi=\left\{\begin{array}{c}
y-\frac{n}{m} x+w \geq 0 ,\\
y-\frac{n}{m} x-w \leq 0 ,\\
0 \leq x \leq m ,\\
0 \leq y \leq n .
\end{array}\right.
\end{equation}

\textbf{Lower Bounding Constraints.} When sequence $L=\left\{L_{1}, L_{2}, L_{3}, \cdots, L_{N}\right\}$ with a total number of $N$ matches the template sequence $Q$, it is necessary to filter out sequences that are not similar to $Q$ in sequence $L$. We use LB\underline{~~}KIM\cite{label23} lower bound function and denote the start element, end element, maximum element, and minimum element of the sequence $L_{i}$ as  $f_{start}, f_{end}, f_{greatest}, f_{smallest}$, and the elements at the corresponding position of the template sequence $Q$ are denoted as $g_{start}, g_{end}, g_{greatest}, g_{smallest}$. The lower bound distance function $D_{tw-lb}(L_{i},Q)$ can be obtained, which is defined as follows:
\begin{equation}
\begin{split}
    D_{tw\underline{~~}lb}(L_{i},Q)=max\{|f_{start}-g_{start}|,|f_{end}-g_{end}|,\\|f_{greatest}-g_{greatest}|, |f_{smallest}-g_{smallest}|\}.
\end{split}
\end{equation}

According to the lower bound theorem\cite{label23}, $DTW(L_{i},Q)\;
\geq\;D_{tw\underline{~~}lb}(L_{i},Q)$. Assuming that there is a similarity threshold $\epsilon$, if $D_{tw\underline{~~}lb}
(L_{i},Q)\;\geq\;\epsilon$, then $DTW(L_{i},Q)\;\geq\; \epsilon$. Thus, we only need to calculate the value of the lower bound function between sequences, and compare it with $\epsilon$ to filter out some dissimilar sequences to greatly reducing the computational complexity of the algorithm.

 \textbf{Early Abandon}\cite{label24}. The lower bound function can only roughly filter out dissimilar sequences, but the remaining sequences may not have high similarity to the template sequence. Our goal is to find sequences with high similarity to $Q$, so we need to stop the sequence matching between dissimilar sequences as soon as possible. Denote the threshold of the cumulative distance of matching between sequences as $\upsilon$. In the $DTW$ path search process, if the cumulative distance between the two sequences $D\;\geq \;\upsilon$, then matching is terminated.
 \subsubsection{Analysis of $DTW$ computational complexity}{}
 \textbf{Assume.} the template sequence $Q$ with length $n$; sequence set $L=\{{L_{1},L_{2},}$ ${L_{3},\cdots,L_{N}}\}$, in which the length of sequence $L_{i}$ is $m$. Then computational complexity of $DTW(L_{i},Q)$ can be expressed as $O(m\times n)$, and the computational cost can be approximately expressed as $m\times n$.
 
 We analyze the influence of the optimization conditions in Part.4.3.1 on the calculation cost of the original $DTW$ one by one. Denote $Cost (DTW_{GC})$, $Cost (DTW_{LB})$, $Cost (DTW_{EA})$ as the calculation cost of global constraint, lower bound function, and early stop impact on the original $DTW$. Then the following formula holds:
 \begin{equation}
     Cost(DTW)=N\times m \times n ,
 \end{equation}
\vspace{-0.7cm}
 \begin{equation}
     Cost(DTW_{GC})=N\times(m\times n-S_{out}) ,
 \end{equation}
 \vspace{-0.5cm}
\begin{equation}
    Cost(DTW_{LB})=(N-V)\times m\times n ,
\end{equation}
\vspace{-0.5cm}
\begin{equation}
    Cost(DTW_{EA})=N\times m\times n\times k ,
\end{equation}
$S_{out}$ in Eq.(4.5) is the element that does not belong to the warping window $\chi$; $V$ in Eq.(4.6) is the sequence filtered by lower bound function $D_{tw\underline{~~}lb}$, and $V \leq N$; in Eq.(4.7), $k\in[0,1]$ . Obviously, $Cost(DTW_{GC})$, $Cost(DTW_{LB})$ and $Cost(DTW_{EA})$ are all smaller than $Cost(DTW)$. These optimization conditions are independent of each other, it is easy to get the following conclusion that when these optimization conditions are combined, the calculation cost of $DTW$ will be less than the cost of $DTW$ caused by any single optimization condition. We can further infer that the cost of our method is:
\begin{equation}
    Cost(DTW_{GC+LB+EA})=(N-V) \times (m \times n-S_{out}) \times k.
\end{equation}
Obviously, it’s clear that the cost of our method satisfies:
\begin{equation}
C \in \{Cost(DTW_{GC}), Cost(DTW_{LB}), Cost(DTW_{EA})\},
\end{equation}
\vspace{-0.6cm}
\begin{equation}
Cost(DTW_{GC+LB+EA})\leq C \leq Cost(DTW).
\end{equation}

\section{Results and Analysis}
Our experiment is divided into two parts, corresponding to the two stages of our method. Specifically, Part.5.1 is the performance experiment of the improved estimation model HigherHRNetXt; Part.5.2 is the performance experiment of the entire Pose-${DTW}$ in the context of two-factor variables. Part.5.3 is the analysis of the influence of the main hyperparameters during the Pose-${DTW}$ experiment on the results of 5.2 experiment.
\subsection{HigherHRNetXt}
\textbf{Dataset.} The COCO dataset contains more than 200,000 images and 250,000 pedestrians marked with 17 key points. HigherHRNetXt is trained on the COCO train2017 dataset and evaluated on the COCO val2017 and test-dev2017 dataset. The training set contains 57,000 images. The validation set and testset contain 5,000 and 20,000 images respectively.

\textbf{Training.} We use Adam optimization algorithm to learn network weights. The hardware facilities are 4 sets of 2080Ti, each with 8-batch size. The basic learning rate is 0.001. Data augmentation includes: random rotation, random translation. Image size scaling to 512$\times$512, training 300 epochs. The deep learning framework is Pytorch1.5 and using ImageNet pre-training weights to initialize the network.
\begin{table}[H]
\centering
\caption{$AP^{0.5:0.95}$, 1G FLOPs is equal to $10^{9}$ FLOPs per sec.}

\setlength{\tabcolsep}{0.8 mm}   
\begin{tabular}{ccccccc}
\hline
Method              & Datasets     & Backbone     & \#Params & GFLOPs & AP   \\ \hline
HigherHRNet         & val2017      & HRNet-W32    & 28.6M    & 47.9   & 67.1 \\
HigherHRNetXt(ours) & val2017      & HRNetXt-W32  & 24.3M    & 45.5   & 67.9 \\ \hline
HigherHRNet         & test-dev2017 & -            & -        & -      & 66.4 \\
HigherHRNetXt(ours) & test-dev2017 & -            & -        & -      & 66.9 \\ \hline
\end{tabular}
\end{table}
\textbf{Test.} The test image size is scaled to 512${\times}$512 without additional data augmentation and the test scale is single. Hardware: 2080Ti. Software: Pytorch 1.5. After adding the ResNeXt module, HigherHRNet has reduced the number of parameters by 15\%, FLOPS decreased by 5\%, and performance has been improved. The average accuracy (AP) on the COCO val2017 and COCO test2017 datasets reached 67.9\% and 66.9\% with improvement of 0.8\% and 0.5\%, respectively. The results are shown in Table 5.1. Compared with the existing bottom-up state-of-art open source pose estimation models, the results are shown in Table 5.2.
\begin{table}[]
\centering

\caption{Test results of some bottom-up state-of-art pose estimation models on COCO test-dev2017. Some results refer to \cite{label6}.}
\setlength{\tabcolsep}{1.2 mm}   
\begin{tabular}{cccccc}
\hline
Method              & Backbone    & InputSize & \#Params & GFLOPs & AP   \\ \hline
OpenPose            & -           & -         & -        & -      & 61.8 \\
Hourglass           & Hourglass   & 512       & 277.8M   & 206.9  & 56.6 \\
PersonLab           & ResNet-152  & 1401      & 68.7M    & 405.5  & 66.5 \\
PifPaf              & -           & -         & -        & -      & 66.7 \\
Bottom-up HRNet     & HRNet-W32   & 512       & 28.5M    & 38.9   & 64.1 \\
HigherHRNet         & HRNet-w32   & 512       & 28.6M    & 47.9   & 66.4 \\
HigherHRNetXt(ours) & HRNetXt-w32 & 512       & 24.3M    & 45.5   & 66.9 \\ \hline
\end{tabular}
\end{table}
\subsection{Pose-${DTW}$}
\textbf{Dataset.} At present, there is no open-world re-ID dataset containing multiple factors. To test the effectiveness of our method, we provide a micro re-ID testset including pedestrian changing and cross-modality two-factor variables, which contains 41 pedestrians, and each pedestrian changes two sets of clothes with different colors (each pedestrian with 40 consecutive frame images). At the same time, the AlignGan\cite{label22} model is used to convert RGB images to infrared radiation (IR) images. A sample of the dataset is shown in Fig. 5.1.
\begin{figure}[H]
\setlength{\abovecaptionskip}{-0.2cm}   
\setlength{\belowcaptionskip}{-0.5cm}    
\begin{center}
\includegraphics[width=11cm,height=5.5cm]{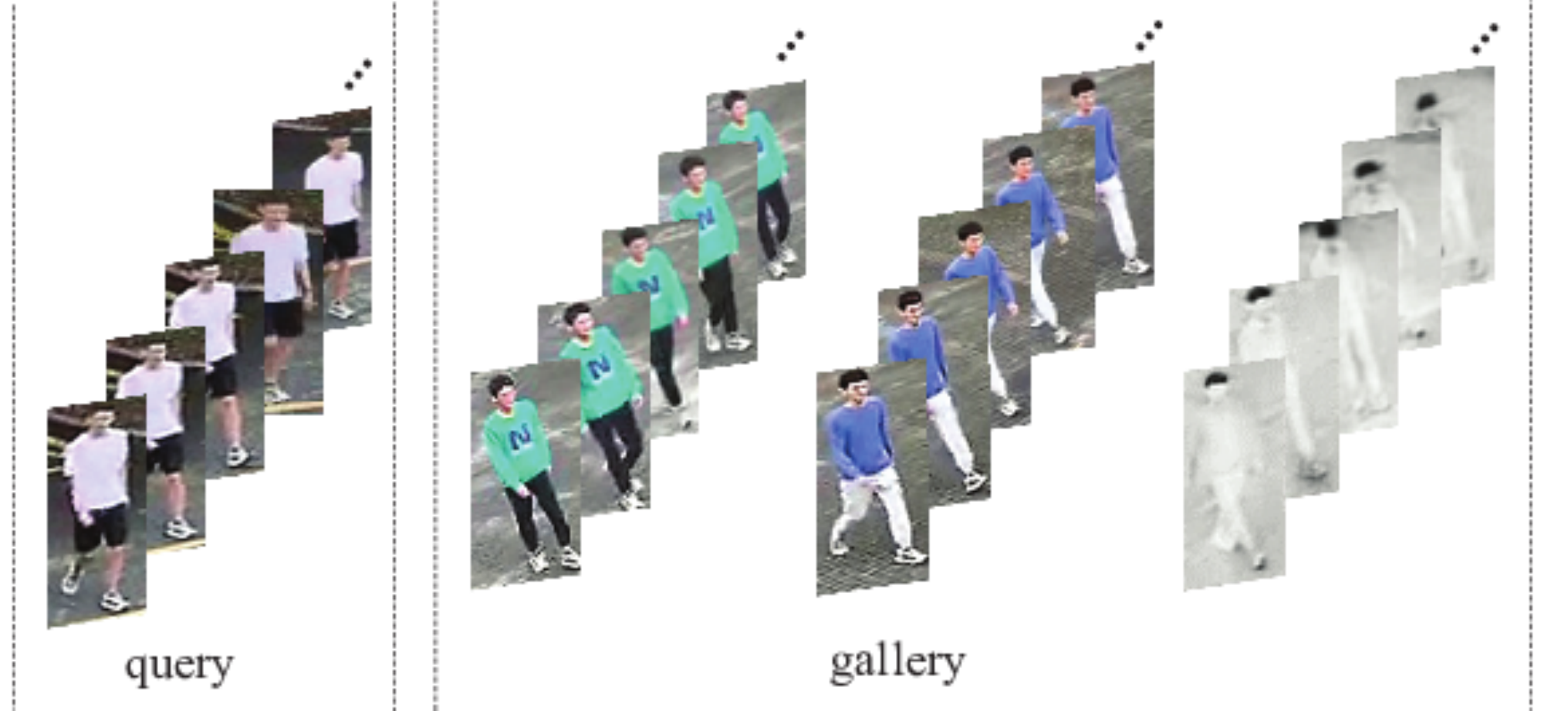} 
\end{center}
\caption{A pedestrian’s image data included in our dataset}
\label{figa10}
\end{figure}

\textbf{Evaluation Standard.} The evaluation of the re-ID model is based on Cumulative Matching Characteristics(CMC) and mean Average Precision(mAP).

\textbf{Test.} The key points of pedestrian skeleton do not include facial key points. The input images are normalized. Hyperparameter settings: $\upsilon =8$, $w =30$, $\epsilon =0.8$. Hardware facilities: 4 sets of 2080Ti. Deep learning framework: Pytorch 1.5. Our method is evaluated on the micro testset together with the existing open source video-based state-of-art re-ID models. The results are shown in Table 5.3. In the context of two-factor variables of pedestrian clothing and cross-modality, our method achieves Rank-1: 60.9\% and mAP: 49.2\% on the testset, which both surpass other state-of-art re-ID models.

\begin{table}[]
\centering
\caption{Performance comparison between our method and the existing open source video-based state-of-art re-ID models on the micro testset.}
\setlength{\tabcolsep}{3.8 mm}   
\begin{tabular}{ccccc}
\hline
\textbf{Methods}         & Source                        & Rank-1        & Rank-5        & mAP           \\ \hline
ETAP-Net\cite{label25}       & CVPR2018                      & 36.6          & 68.2          & 38.4          \\
TriNet\cite{label26}         & arXiv2017                     & 34.1          & 61.0          & 34.4          \\
RRU+STIM\cite{label27}       & AAAI2019                      & 41.5          & 63.4          & 33.2          \\
STE-NVAN\cite{label28}       & BMVC2019                      & 46.3          & 75.5          & 42.3          \\
Part-Aligned\cite{label29}   & ECCV2018                      & 48.8          & 73.2          & 42.6          \\
RevisitTempPool\cite{label30} & BMVC2019                      & 43.9          & 70.7          & 40.1          \\
\textbf{Pose-${DTW}$(ours)}  & \multicolumn{1}{l}{\textbf{}} & \textbf{60.9} & \textbf{78.1} & \textbf{49.2} \\ \hline
\end{tabular}
\end{table}

\subsection{Hyperparameter analysis}
The main hyperparameters in our method are concentrated in Stage.2, including the cumulative distance threshold $\upsilon$, the warping window width $w$, and the lower bound threshold $\epsilon$. Their values will affect the matching accuracy of the time series features of $L_{i}$ and $Q$. If the value of these hyperparameters is too small, a lot of information will be lost; if the value is too large, the calculation complexity of the algorithm cannot be reduced. Therefore, we conducted some experiments independently to determine the values of these hyperparameters. The experimental results are shown in Fig. 5.2. The initial value of $w$ is 10, the end value is 40, and the step length is 10; the initial value of $\upsilon$ is 3, the end value is 8, and the step length is 1; the initial value of $\epsilon$ is 0.2, the end value is 0.8, and the step length is 0.2.
\begin{figure}[H] 
\centering 
\subfigure[]{
\label{fig6a}
\includegraphics[width=3.7cm,height=3.2cm]{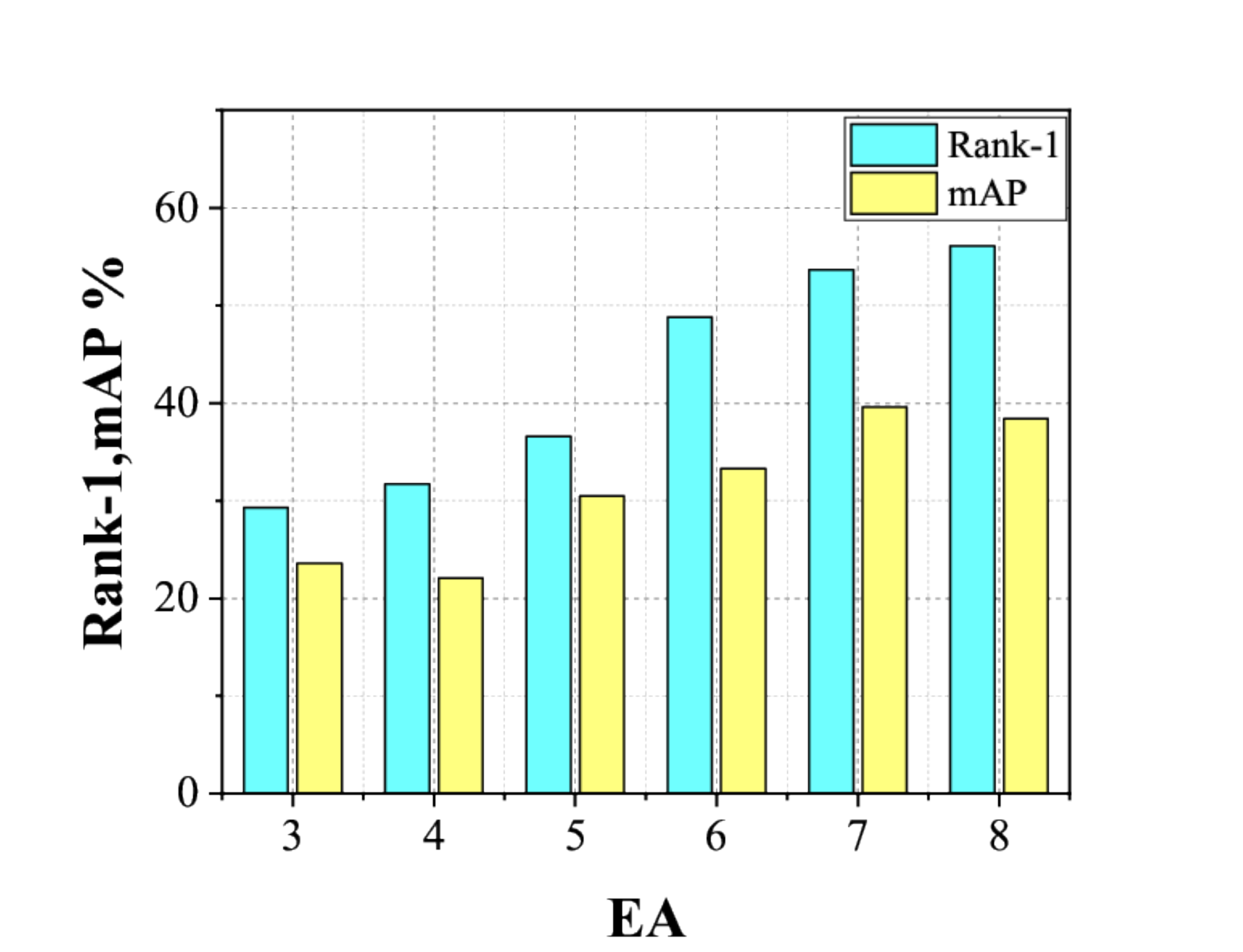}
}
\subfigure[]{
\includegraphics[width=3.7cm,height=3.2cm]{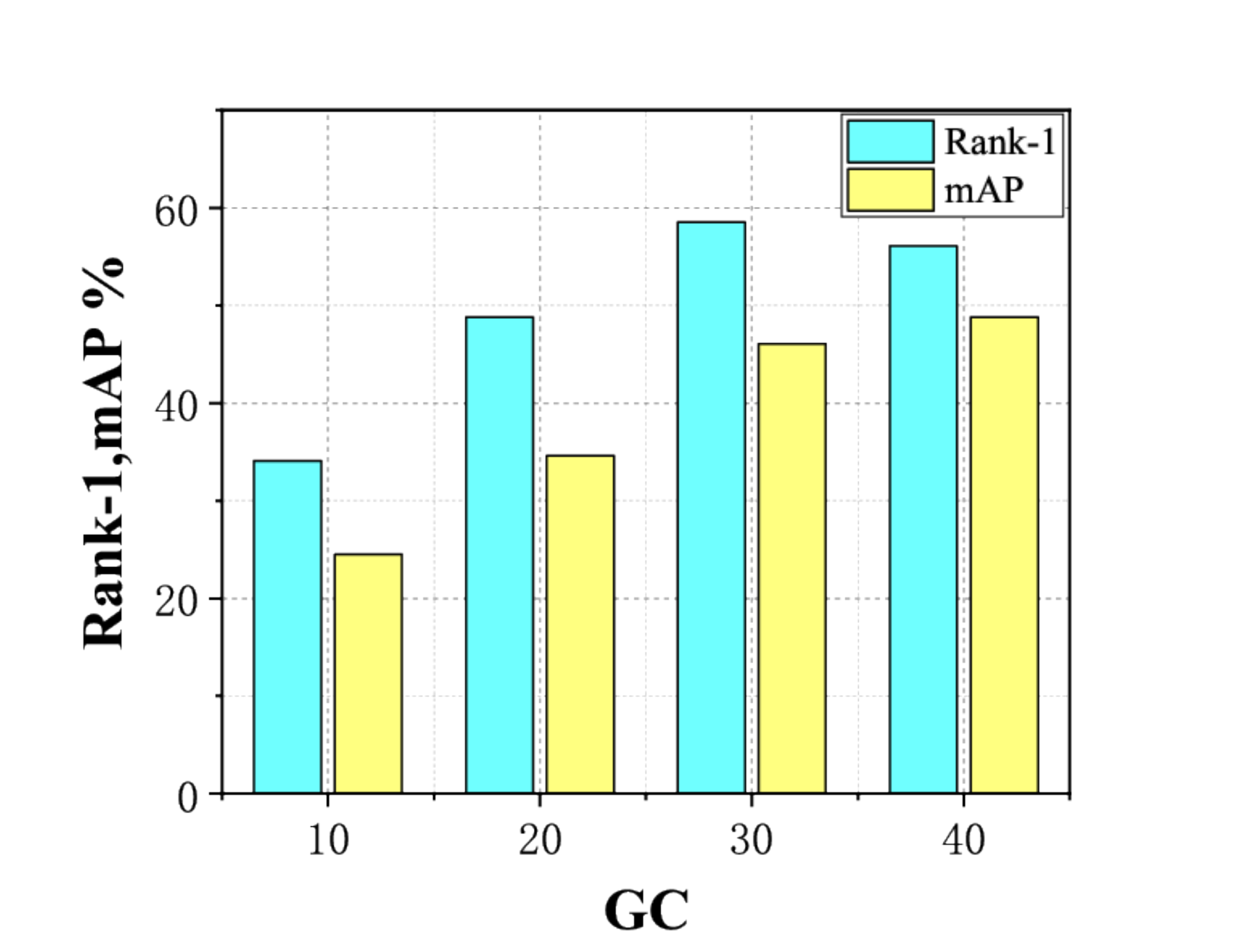}
\label{fig6b}
} 
\subfigure[]{
\label{fig6c}
\includegraphics[width=3.7cm,height=3.2cm]{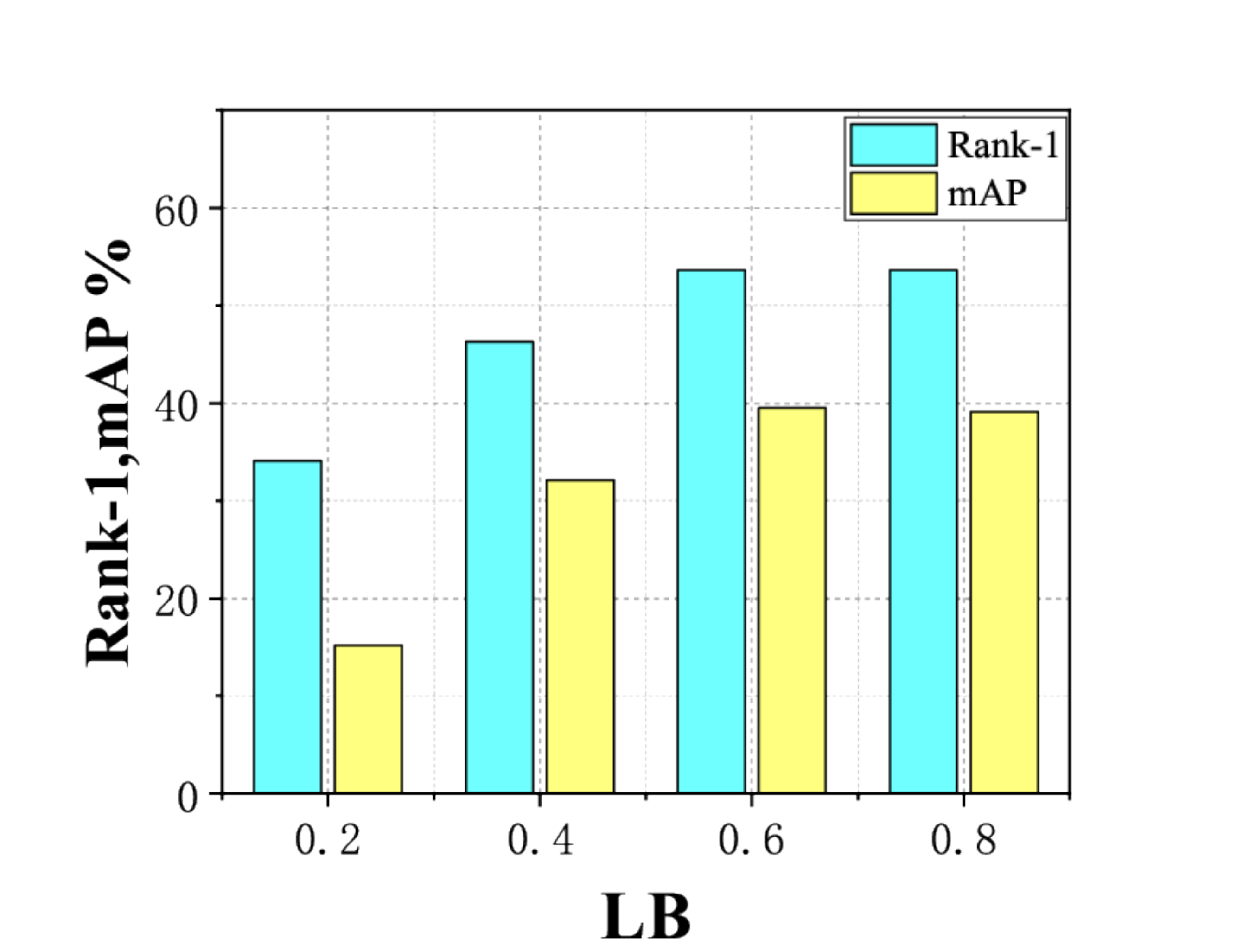}
}
\caption{(a), (b), (c) are the effects of hyperparameters, $\upsilon$, $w$, and $\epsilon$ on the Rank-1 and mAP of Pose-${DTW}$ on the micro testset, respectively. The experiments of Hyperparameters are carried out independently of each other, the value of $\upsilon$ is determined by the Pose-$DTW_{GC}$ experiment, at the same time the values of $w$ and $\epsilon$ are empty. The values of $w$ and $\epsilon$ are also determined in this way.}
\end{figure}

It can be seen from Fig. 5.2 that the suitable value of $w$ is 30; the suitable value set of $\upsilon$ is \{6,7,8\}; the suitable value set of $\epsilon$ is \{0.6,0.8\}. Theoretically the larger the value of these hyperparameters, the higher the complexity of the algorithm. Therefore, when the experimental results of Rank-1 and mAP are similar, the smaller values should be selected as much as possible. Moreover, the hyperparameters $\upsilon$, $w$ , and $\epsilon$ are mutually independent, so their respective optimal values are the optimal values of the combination. For this testset, $\upsilon$, $w$ , and $\epsilon$ are determined to be 7, 30, and 0.6 respectively.

However, we randomly combined the values of these hyperparameters in the experiment with 6 groups:
$\{w=30, \upsilon=6, \epsilon=0.6\}$, $\{w=30, \upsilon=6, \epsilon=0.8\}$, $\{w=30, \upsilon=7, \epsilon=0.6\}$, $\{w=30, \upsilon=7, \epsilon=0.8\}$, $\{w=30, \upsilon=8, \epsilon=0.6\}$, and $\{w=30, \upsilon=8, \epsilon=0.8\}$. Results are shown in Fig. 5.3. The experimental results shows that hyperparameters combination $\{w=30, \upsilon=7, \epsilon=0.8\}$ are better than $\{w=30, \upsilon=7, \epsilon=0.6\}$. We theoretically analyze that $\upsilon$, $w$ , and $\epsilon$ are independent of each other, but there is a slight deviation between the experimental results and the theoretical results. We believe that it is caused by the testset being too small.

\begin{figure}[H]
\setlength{\abovecaptionskip}{-0.2cm}   
\setlength{\belowcaptionskip}{-0.5cm}    
\begin{center}
\includegraphics[width=8cm,height=5cm]{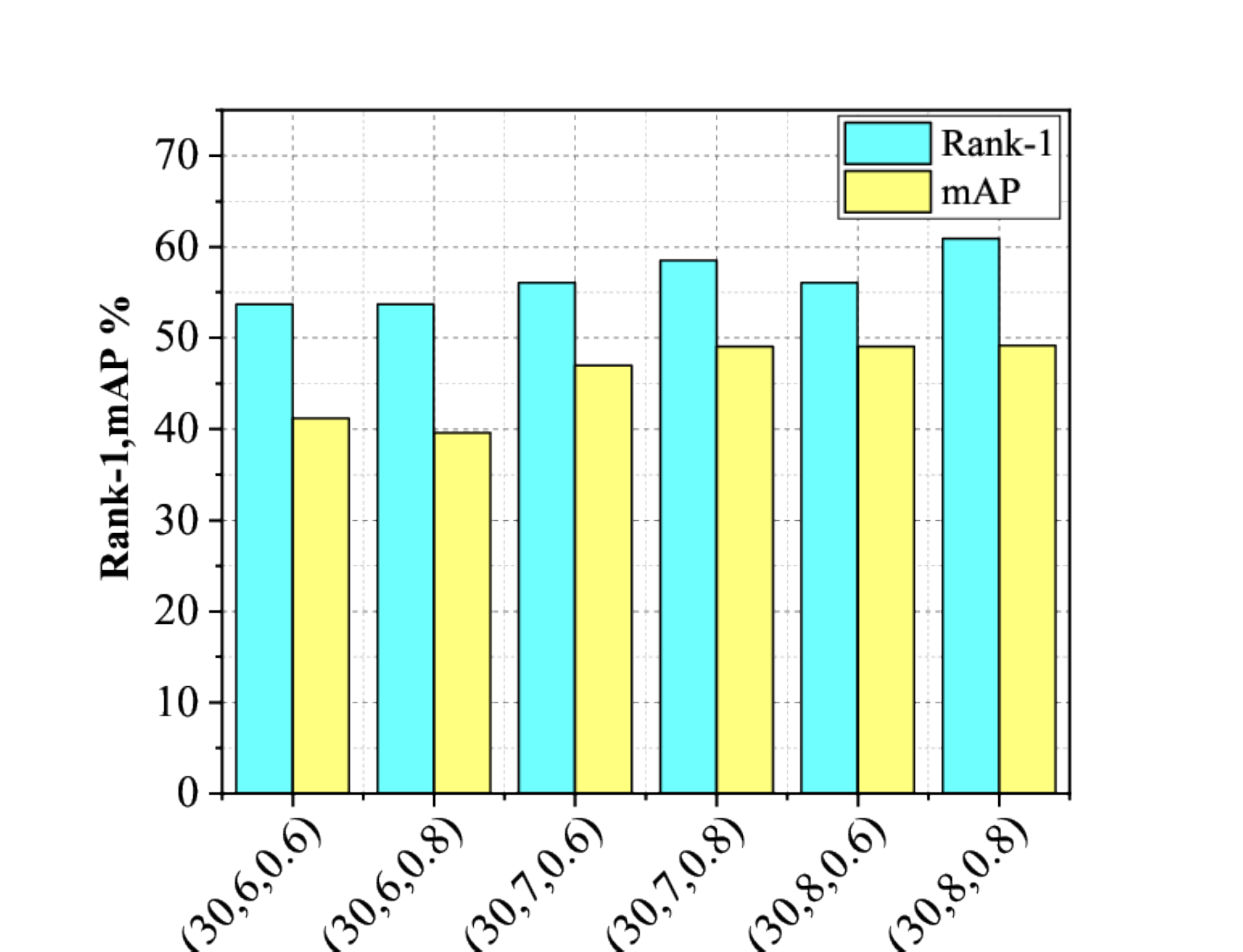} 
\end{center}
\caption{The effect of hyperparameters combination on the performance of Pose-${DTW}$ on the micro testset}
\label{figa12}
\end{figure}

\section{Conclusions}
This paper analyzes the reasons for the poor availability of existing re-ID research in open-world, then analyzes and refines the existing re-ID research problems under the research assumptions. Farther more, we propose a method of combining deep learning and traditional algorithms methods to solve this problem, and the optimization strategies are adopted to optimize the algorithm. Compared with the existing re-ID research, our method proposed in this paper has three advantages. Moreover, it overcomes the problem of lack of open-world re-ID dataset and improves the usability of re-ID in open-world. To the best of our knowledge, we are the first to conduct experiments on re-ID under the double-factor background of pedestrian clothing and cross-modality.

Our method is mainly oriented, hoping to further stimulate more research on re-ID in open-world. There are still shortcomings of our model. For example, Pose-${DTW}$ almost completely discards the color features information of pedestrians, which causes a waste of feature information, especially when the information obtained is limited. So, we guess that combining our method with the existing re-ID model will have better performance. However, limited by our small test dataset, it seems to be difficult for a further experiment, which is our next work.

\section*{Acknowledgments}
This research is supported by the Zhejiang Provincial Public Welfware Technological Project of China under Grant No.LGF20F020007 and the National Natural Science Foundation of China under Grant No.11801511.
\bibliography{mybibfile}

\end{document}